\documentclass[runningheads]{llncs}
\usepackage[T1]{fontenc}
\usepackage{graphicx}
\usepackage{subfig}
\usepackage{booktabs}  
\usepackage{hyperref}
\begin{document}
\title{Exploring Gender Differences in Chronic Pain Discussions on Reddit}
\titlerunning{Gender differences in Chronic Pain Discussions}
%
\author{Ancita Maria Andrade\inst{1} \and
Tanvi Banerjee\inst{1} \and
Ramakrishna Mundugar\inst{2}}
%

%
\institute{
Wright State University, Dayton, OH 45435, USA\\
\email{\{andrade.12, tanvi.banerjee\}@wright.edu} 
\and
Manipal Institute of Technology, Manipal, Karnataka 576104, India\\
\email{ramakrishna.m@manipal.edu}
}
\maketitle              
\begin{abstract}
Pain is an inherent part of human existence, manifesting as both physical and emotional experiences, and can be categorized as either acute or chronic. Over the years, extensive research has been conducted to understand the causes of pain and explore potential treatments, with contributions from various scientific disciplines. However, earlier studies often overlooked the role of gender in pain experiences. In this study, we utilized Natural Language Processing (NLP) to analyze and gain deeper insights into individuals' pain experiences, with a particular focus on gender differences. We successfully classified posts into male and female corpora using the Hidden Attribute Model-Convolutional Neural Network (HAM-CNN), achieving an F1 score of 0.86 by aggregating posts based on usernames. Our analysis revealed linguistic differences between genders, with female posts tending to be more emotionally focused. Additionally, the study highlighted that conditions such as migraine and sinusitis are more prevalent among females and explored how pain medication affects individuals differently based on gender.
\end{abstract}
\section{Introduction}

Social media (SM) has emerged as a transformative tool, reshaping communication and interaction in the digital age. According to Wikipedia, SM refers to interactive technologies that facilitate the creation, sharing, and aggregation of content—such as ideas, interests, and other forms of expression—within virtual communities and networks \cite{b41}. Over the past two decades, SM platforms have experienced unprecedented growth, driven by advancements in internet access, mobile technology, and the widespread adoption of smartphones. As of January 2025, there are 5.24 billion SM users, constituting 63.9\% of the global population \cite{bstat}, fostering collaboration and networking at a global scale \cite{b44}.


Numerous studies have explored both the nature of social media (SM) and its application in research, with over 110,000 publications featuring the term "social media" in their titles as of January 2020 \cite{b42}. Beyond its numerous other uses, social media now plays a crucial role in public health by empowering health promotion, education campaigns, intervention deployment, disease surveillance systems, and research activities \cite{b45}.

One critical public health issue that has gained attention in this context is chronic pain—a long-standing challenge affecting millions worldwide. In general, 1 in 10 people are diagnosed with chronic pain conditions annually \cite{b4}. Researchers have been working on exploring how pain varies across various demographics including age, gender, ethnicity, economic status, etc. for many years. Several works have shown that men and women experience pain differently. It is observed that women are more sensitive to pain and make up a large population that experiences pain \cite{b3}. The International Association for the Study of Pain (IASP) designated 2024 as the Global Year for Sex and Gender Disparities in Pain to highlight differences in pain perception and treatment, and address related gaps in research and care (\url{https://www.iasp-pain.org/advocacy/global-year/sex-and-gender-disparities-in-pain/}).

The perception of pain is subjective and healthcare professionals assess patients according to the description they provide. Several studies have shown that specific linguistic markers exist for areas such as pain catastrophizing, mood, as well as diagnostic categories \cite{b5}. However, language is not neutral; it often reflects and reinforces social norms and biases, including those related to gender. Using the Social Categories and Stereotype Communication (SCSC) framework \cite{b6}, we can understand how social categories influence the way stereotypes are conveyed and reinforced within communication, in turn influencing impressions of people and groups. It is crucial to take into account all aspects of language and stereotypes in this research, particularly the idea that men are often less willing to share personal experiences. This can help gain a deeper understanding of the common underlying issues faced by individuals in pain.

The role of gender stereotypes makes SM an essential starting point for our research.   Reddit is one of the popular SM platforms and liked by many of the users (redditors) for its ecosystem of various communities for different topics. In one of the subreddits, r/CasualConversation, a question was asked about why people (redditors) share their personal stories on Reddit, the most voted answer was due to reasons such as anonymity and wanting to share with the people who would not judge. Researchers using Reddit data have found empirical evidence that members openly discuss and share informational support on potentially stigmatized issues \cite{b23}. Subreddit moderators oversee the communities, ensuring that users post content relevant to the discussion topics. This is why Reddit makes an ideal platform for our research.

Our study aimed to capture patient-driven narratives and explore gender-related differences in chronic pain discussions through peer-to-peer communication. It provides valuable firsthand insights that may not always be captured in traditional clinical settings. Gender is a social, psychological, and cultural construct that defines the social characteristics of men and women. It differs from sex, which primarily refers to biological and physiological attributes \cite{b46}. In this study, we focused on gender identity and limited our analysis to a binary classification of male and female as non-binary users make up only 1.3\% of the total reddit users \cite{b47}. 


Ethical considerations are essential in the use of social media data for research purposes. In accordance with Reddit’s User Agreement, users acknowledge that their comments are publicly accessible and may be retrieved through the platform’s API. This study complied with Reddit’s terms of use and employed measures to safeguard user anonymity by excluding identifiable information such as usernames. Furthermore, only partial excerpts from posts were presented to ensure that individual users could not be identified.

We explored the following questions in this work.
\begin{enumerate}
  \item How effectively can supervised machine learning techniques predict an author's gender based on features extracted from their Reddit posts or comments?
  \item To what extent can we use data-driven techniques to elicit common ways in which men and women differ in expressing their pain?
  \item How effective are techniques such as topic modeling in identifying gender-based differences in discussions related to chronic conditions and the experiences shared by individuals?
  \item What are the most important differences in patterns of medication usage between men and women using Named Entity Recognition (NER) and sentiment analysis?
\end{enumerate}

\section{Related Work}
The basis for our study rests on the results of gender prediction. For any supervised machine learning model, there has to be a labeled dataset. RedDust is the collection of user traits collected by the authors \cite{b1} using the assertions made by the redditors. The dataset includes user attributes such as gender, age, family status, profession and hobby. The work also discusses various predictive models for various attributes with Hidden Attribute Model-Convolution Neural Networks (HAM-CNN) being the best for gender prediction.

Another significant social media platform is Twitter, which has been extensively studied for author profiling. Most approaches utilize supervised learning algorithms, ranging from simple methods like logistic regression and SVM classifiers to advanced deep learning techniques such as GRUs and other deep neural networks. One of the notable researches \cite{b9} includes classifying tweets, incorporating images.

Analyzing Reddit data related to chronic pain is not novel. For example, the authors of a study \cite{b8} examined 12 subreddits focused on chronic pain, identifying the primary concerns discussed within each community. Some subreddits addressed diverse topics, others shared overlapping concerns, such as r/backpain and r/sciatica, which both prominently discussed issues related to back pain and related symptoms. They also analyzed the demographics of users who disclosed their gender and age. Among these users, 23.53\% reported their gender, with 55.51\% identifying as female and 44.49\% as male.


Another study examined gender differences in pain expression by asking 201 participants to describe a past painful experience. Results showed that women tended to use more emotional and descriptive language, focusing on pain intensity and affective aspects, while men employed more objective terms, emphasizing physical location and resolution \cite{b2}.
\section{Methods}
\begin{figure*}[htbp]
\centerline{\includegraphics[width=\textwidth]{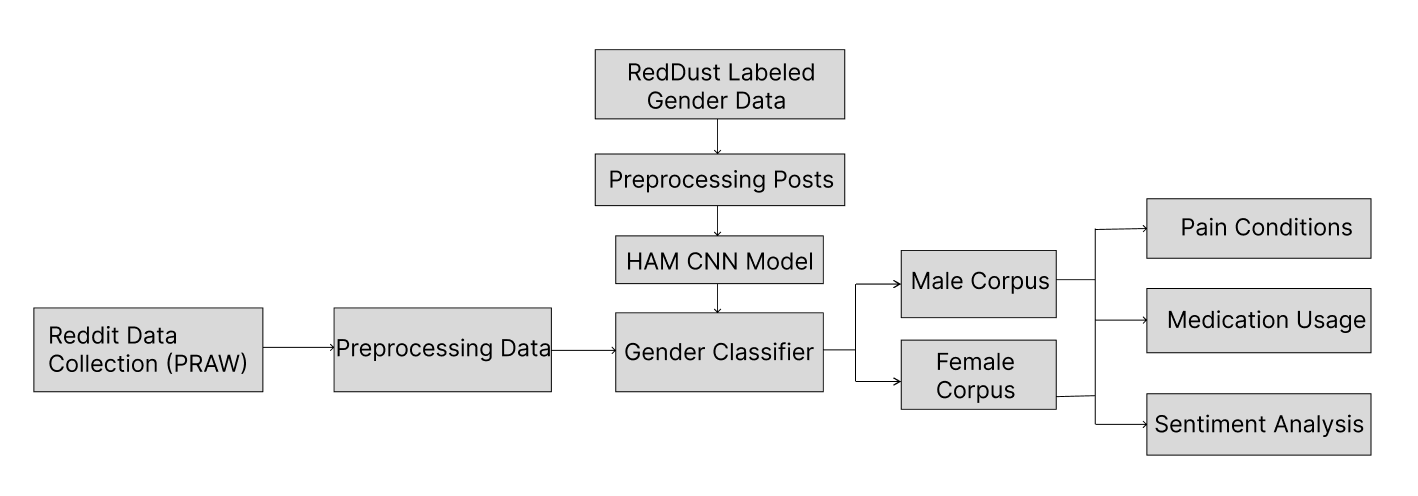}}
\caption{Pipeline of the project.}
\label{fig:pipeline}
\end{figure*}

We employed various Natural Language Processing (NLP) techniques to analyze patterns in posts and comments made by male and female users. The overall workflow of the project is illustrated in the figure \ref{fig:pipeline}.

\subsection{Data Collection}
To develop the gender classification model, we leveraged the existing Reddit user attribute dataset, RedDust\cite{b1}. Additionally, we collected data from Reddit using the publicly available Python Reddit API Wrapper (PRAW) for research purposes. We manually selected 36 subreddits relevant to chronic pain discussions. These included general subreddits such as r/ChronicPain and r/Pain, as well as those focused on specific chronic conditions, including r/Migraine and r/Sinusitis. Data was extracted up to March 31, 2025. Data included 28,067 posts along with their comments from 58k redditors. 

\subsection{Data Cleaning}
We included only posts and comments containing at least three words and excluded those not written in English. Entries authored by moderators or bots (e.g., 'Automoderator', 'cfs-modteam') were removed from the dataset. We further refined the dataset by removing duplicate posts from the same author, as there were instances where Redditors had shared identical content across multiple subreddits.

\subsection{Gender Classification}
To address RQ1, we employed the RedDust dataset, with a particular focus on Reddit users and their gender identities. The dataset contains 2.49 million posts from around 54.8k users. For initial model training, each post or comment was treated as an independent data point (i.e., post-level classification). This approach was chosen because many Reddit users utilize "throwaway" accounts and may post only once.

We experimented with both traditional and deep learning classification methods, beginning with a baseline Logistic Regression model and advancing to more sophisticated architectures such as the bi-directional Gated Recurrent Unit (GRU) model, as described in \cite{b9}, and the HAM-CNN model introduced in \cite{b1}. The performance of these models is summarized in Table \ref{tab:gender_classifier_models}. To mitigate class imbalance, we applied undersampling to the majority class (Female) and used a stratified train-test split.

Among the post-level models, HAM-CNN achieved the highest performance with an F1-score of 0.67. As shown in \cite{b16}, models generally achieve significantly better performance when trained at the user level—by aggregating all posts from each user and treating the user as a single data point—often resulting in nearly double the F1-score compared to post-level training. Motivated by this, we aggregated posts and comments by user and retrained the HAM-CNN model using user-level data. This approach significantly outperformed the post-level models, yielding an F1-score of \textbf{0.86} and an AUC of \textbf{0.94}.

To classify the Reddit data we collected, we organized the data based on usernames. Using the trained HAM-CNN model, we predicted the gender of Reddit users. To ensure unbiased predictions, we only included users with high confidence scores—greater than 0.75 for females and less than 0.25 for males. This filtering process resulted in a final set of approximately 43K Redditors.

\begin{table}[htbp]
\caption{Metrics for the gender classifier for different models}\label{tab:gender_classifier_models}
\centering
\begin{tabular}{|l|l|p{2cm}|p{2cm}|p{2cm}|}
\hline
\bfseries Model & \bfseries Gender & \bfseries Precision & \bfseries Recall & \bfseries F1 Score \\ 
\hline
Logistic Regression
(Post Level)& Male    & 0.64 & 0.66 & 0.65 \\ 
\cline{2-5}
                    & Female  & 0.65 & 0.63 & 0.64 \\ 
\hline
Bi GRU (Post Level)    & Male    & 0.67 & 0.68 & 0.67 \\ 
\cline{2-5}
                    & Female  & 0.67 & 0.66 & 0.67 \\ 
\hline
HAM-CNN  (Post Level)           & Male    & 0.66 & 0.72 & 0.69 \\ 
\cline{2-5}
                    & Female  & 0.69 & 0.63 & 0.66 \\ 
\hline
\textbf{HAM-CNN 
(User Level)}         & Male    & \textbf{0.86} & \textbf{0.86} & \textbf{0.86} \\ 
\cline{2-5}
                    & Female  & \textbf{0.86} & \textbf{0.87} & \textbf{0.87} \\ 
\hline
\end{tabular}
\end{table}

\subsection{Gender Differences in Posting Behavior}
As a necessary precursor to NLP analysis, we first examined the posting behavior of male and female users on Reddit. Table \ref{tab:EDA} represents various feature statistics. We found that the number of male Redditors is higher than that of female Redditors in chronic pain-related subreddits. This pattern is consistent with the overall gender distribution observed across Reddit. However, the number of posts made by female users is nearly twice that of male users, indicating higher posting activity among females in chronic pain-related subreddits. This may suggest that female users are more likely to seek social support through online communities compared to their male counterparts \cite{b18}. The average post length and word count indicate that women tend to write more detailed and expressive posts, whereas men are generally more concise and assertive in their communication \cite{b26}. We analyzed the readability of posts using the Flesch Reading Ease score from the textstat Python library. The results showed that male-authored posts had higher readability scores, indicating that they were generally easier to read compared to those written by females. We also assessed lexical diversity, which measures the variety of unique words used in a text. It is typically calculated as the ratio of unique words (types) to the total number of words (tokens). Our analysis revealed that male-authored posts exhibited slightly higher lexical diversity, suggesting a broader vocabulary usage compared to female-authored posts.

\begin{table}[htbp]
    \centering
    \renewcommand{\arraystretch}{1.3} 
    \setlength{\tabcolsep}{12pt} 
    \caption{Feature Statistics}
    \label{tab:EDA}
    \begin{tabular}{|l|r|r|}
        \hline
        \textbf{Textual Features} & \textbf{Male} & \textbf{Female} \\ \hline
        Number of Redditors       & 30,677        & 12,605          \\ \hline
        Number of Posts           & 3,567         & 7,234           \\ \hline
        Avg Number of Comments    & 6.30          & 12.08           \\ \hline
        Avg Post Length (chars)   & 242.81        & 387.88          \\ \hline
        Avg Word Count            & 45.19         & 71.75           \\ \hline
        Avg Flesch Reading Ease   & 73.21         & 72.34           \\ \hline
        Avg Lexical Diversity     & 0.90          & 0.86            \\ \hline
    \end{tabular}
\end{table}

\subsection{Sentiment Analysis}
Redditors use the platform to share everything ranging from personal triumphs to bitter experiences. LIWC (Linguistic Inquiry and Word Count) is a text analysis tool that quantifies linguistic and psychological attributes in written or spoken language by categorizing words into predefined psychological, social, and structural categories. We performed a LIWC analysis (\url{https://www.liwc.app}) on both male and female datasets to explore various affective attributes, such as positive and negative tone, as well as the use of swear words. To gain deeper insights into the sentiments expressed by users, we further analyzed emotional content using Hugging Face Transformers. The emotions considered in this analysis included sadness, joy, anger, fear, surprise, disgust, and neutrality.

\subsection{Topic Modeling}
\begin{figure}[htbp]
\centering
\includegraphics[width=0.6\linewidth]{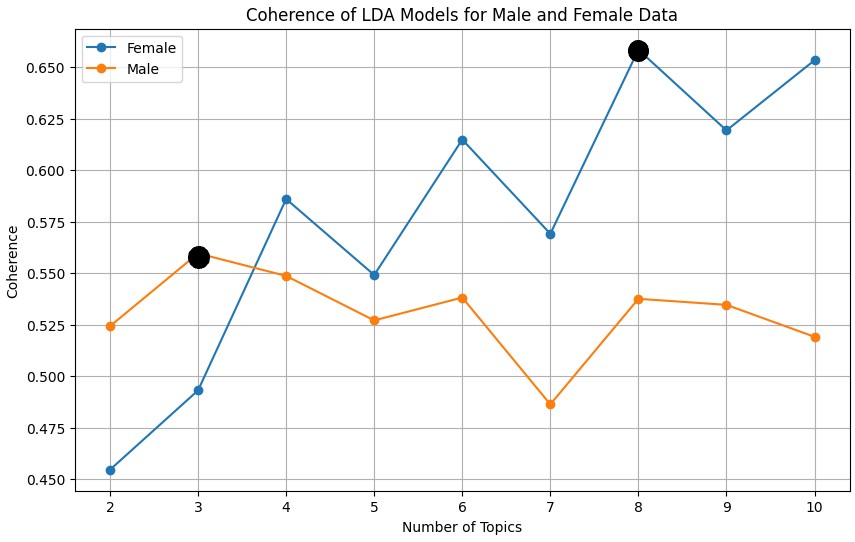} 
\caption{Scores vs. Topics for Male and Female Data}
\label{fig:n_for_both}
\end{figure}

Topic modeling is a key step in NLP, often used as an unsupervised method to uncover hidden themes within textual data. In our study, we applied Latent Dirichlet Allocation (LDA) \cite{b10} separately to the male and female corpora to identify underlying topics. To determine the optimal number of topics, we evaluated the model’s coherence scores across a range of topic numbers (n), from 2 to 10. A higher coherence score indicates better topic interpretability, suggesting that the words within each topic are more semantically coherent and logically related. As shown in Fig. \ref{fig:n_for_both}, the optimal number of topics identified for the female corpus is 8, whereas for the male corpus, it is 3.

\subsection{Drugs Discussion}
Named Entity Recognition (NER) is a NLP technique that detects and classifies specific entities within a text into predefined categories. It identifies and classifies entities like names, locations, or dates using NLP techniques, providing context-aware and structured insights. We utilized the Python library \textbf{Drug Named Entity Recognition (NER)} to extract drug mentions from the text. This tool identifies only high-confidence drug names but does not capture abbreviations or short forms. For instance, in a post where a male user stated, \textit{"If anyone knows of one in KY to help me with my chronic pain I'd appreciate it. Was prescribed oxy and methadone,"} the NER successfully identified ``Methadone'' but missed ``oxy'' (likely referring to oxycodone).

We assumed that any drug mentioned within a post or comment is significant, whether users are discussing their experience using the drug, reporting its ineffectiveness, or seeking opinions about it.
A few examples include:
\begin{itemize}
    \item A male user shared, \textit{"So my ortho doc said they could do cortisone/lidocaine injections,"} where NER correctly extracted ``Cortisone'' and ``Lidocaine.''
    \item A female user mentioned, \textit{"I’ve also had Gabapentin for nerve pain before, but I didn’t find it helpful at all,"} from which NER identified ``Gabapentin.''
\end{itemize}

In the dataset, users mentioned a variety of drugs; however, our analysis focused on those with the highest frequency of mentions, specifically Amitriptyline, Gabapentin, Lidocaine, Morphine, Pregabalin, and Tramadol. Data regarding the specific medical conditions for which these drugs were used or the dosages consumed were excluded, as such information was inconsistently reported by users. Furthermore, the study aimed to address a broad spectrum of chronic conditions rather than restricting the analysis to particular diagnoses.

To evaluate the efficacy of the medications, we employed a rule-based classification approach, categorizing posts into four distinct groups: \textit{effective}, \textit{ineffective}, \textit{side effects}, and \textit{neutral}. For effective posts, we identified keywords such as \textit{``improvement,'' ``helped,'' ``worked,''}, and similar terms. Posts containing phrases like \textit{``didn't work,'' ``waste,'' ``no effect,''} and related expressions were classified as ineffective. To detect side effects, we extracted symptoms referenced from the study by \cite{b32}. The remaining posts, which did not match any of these criteria, were labeled as neutral. We conducted comprehensive descriptive and statistical analysis to evaluate both the therapeutic effects and emotional sentiment expressed regarding medications across different gender groups.

\section{Results and Discussion}
\subsection{Sentiment Analysis}

\begin{figure}[!t]
    \centering
    \subfloat[Comparing the Sentiments across female and male posts \label{fig:Sentiment}]{
        \includegraphics[width=0.48\linewidth]{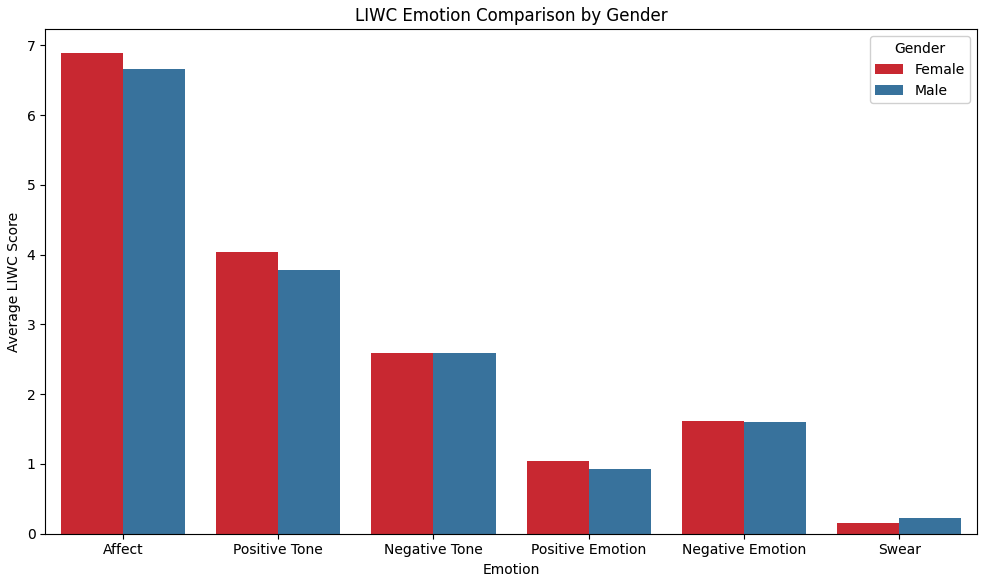}
    }
    \hfill
    \subfloat[Comparing the Emotions across female and male posts\label{fig:Emotions}]{
        \includegraphics[width=0.48\linewidth]{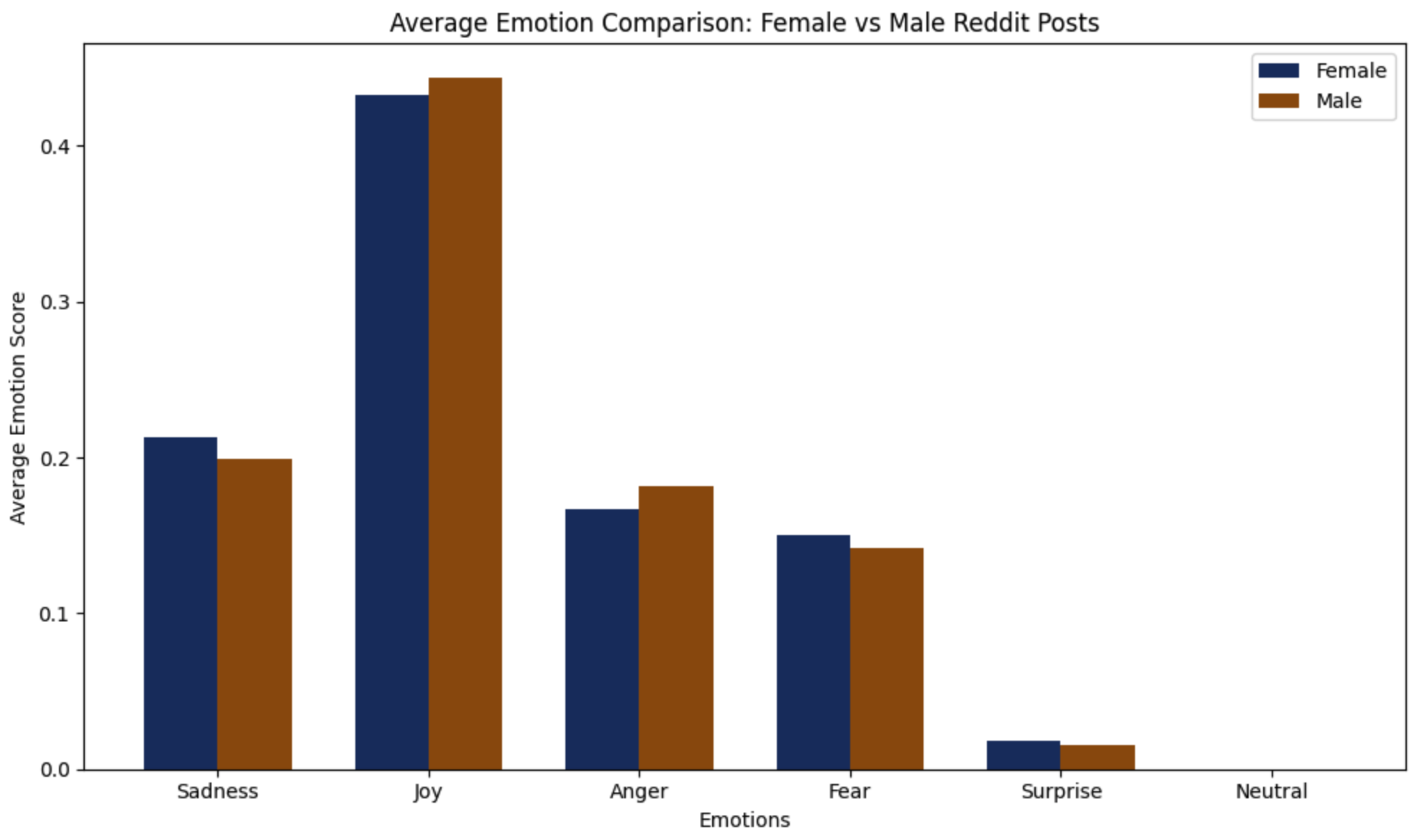}
    }
    \caption{Comparison of topic modeling results by gender}
    \label{fig:gender_comparison}
\end{figure}

We analyzed the LIWC affect features separately for each gender to answer our RQ2. Figure \ref{fig:gender_comparison} shows the sentiment distribution among male and female. Sentiment analysis of female data showed that they tend to share more intensely positive and/or negative experiences than men. While women displayed a broader spectrum of emotional expression. This is corroborated by \cite{b15} in which the author discusses how women tend to have stronger emotional memory due to the interplay of the amygdala (the brain's emotional center) and the hippocampus (the brain's memory center). In contrast, men were found to use more swear words. Recent studies also suggest that females generally have larger volumes of the orbital frontal cortex, which helps regulate anger and aggression triggered by the amygdala. This neurological difference may contribute to the observed gender variation in the use of strong language\cite{b27}.

In analyzing the emotions expressed in discussions about chronic pain, we found that female users exhibited higher levels of fear and sadness. A closer examination of posts with elevated fear scores revealed that these were often centered around discussions of symptoms and side effects of medications. These findings are consistent with those reported in previous research \cite{b28}, which also highlights that females reported higher pain related frustration and fear. In a related study \cite{b29}, the author noted that individuals often express anger either toward themselves or their healthcare providers. Our analysis revealed that male users expressed higher levels of anger compared to females. Their posts frequently described feelings of frustration and irritation directed at themselves and their family members. However, many also expressed a sense of guilt or remorse for feeling this way.
\subsection{Topic Modeling}
\begin{table}[!htb]
\centering
\caption{LDA Topics Extracted from Male Posts}
\label{tab:topic_model_male}
\renewcommand{\arraystretch}{1.3}
\begin{tabular}{@{}c p{4.5cm} p{7.5cm}@{}}
\toprule
\textbf{Topic No} & \textbf{Theme and Associated Words} & \textbf{Example Posts/Comments} \\
\midrule
1 & \textbf{Exercises \& Musculoskeletal Health} \newline nerve, leg, exercise, disk, pt, muscle, stretch, diet, mri, supplement & 
" I've seen several guys focus way to hard on deadlift, and not prioritize core, and get really bad anterior pelvic tilt and way too much curve." \newline
"I wouldn't stop lifting all together, this could actually make things worse" \\
\midrule
2 & \textbf{Social coping} \newline need, doctor, life, sorry, best, post, diagnose, experience, hope, ask &
" I want to manage tolerance as best as possible."  \newline "I suppose it's natural to start seeing the world and life from a very different perspective than most people."\\
\midrule
3 & \textbf{Temporal chronic pain experience} \newline bad, night, work, week, sleep, year, hour, migraine, hour, month & "I had longer period of back pain after lifting about 1 year ago." \newline
"ive been experiencing some weird symptoms that feels like they originate from my upper
stomach (below chest)." \\
\bottomrule
\end{tabular}
\end{table}

In case of topic modeling, we interpreted the topics of LDA from both male and female data. Table \ref{tab:topic_model_male} and Table \ref{tab:topic_model_female} includes the result of LDA. LDA revealed several insights into gender differences, with two major themes standing out: differences in commonly discussed chronic conditions and variations in how individuals express their experiences.

Incase of female corpus, out of 8 topics, 6 were related to various chronic conditions. Vulvodynia, endometriosis, Irritable bowel syndrome (IBS), sinusitis, lower back pain and migraine were the conditions which were highlighted. Previous research has confirmed that these conditions are indeed more prevalent among females than males \cite{b30} \cite{b31}. Topic 3 highlighted how women expressed the impact of chronic pain on their lives, while Topic 7 focused on the theme of social support. The presence of tokens such as "helpful" and "appreciate", "share" reflects the emotional depth in female narratives, suggesting a tendency toward empathy and emotional expression. This supports the idea that women are more inclined to seek social support and adopt emotion-focused psychological coping strategies \cite{b18} \cite{b21} \cite{b22}.

In the male dataset, Topic 1 primarily centered around discussions on physical activity and musculoskeletal health, while Topic 2 highlighted how men cope with chronic pain. Their conversations emphasized physical self-care strategies such as structured strength training, personalized rehabilitation programs, appropriate medication, and active rest to manage various conditions. This pattern suggests that men tend to adopt problem-focused coping techniques and behavioral distraction, prioritizing tangible, solution-oriented approaches over emotional expression. These findings align with previous research indicating that men are more likely to address the physical realities of pain management rather than its emotional dimensions \cite{b2}.
\begin{table}[!htb]
\centering
\caption{LDA Topics Extracted from Female Posts}
\label{tab:topic_model_female}
\renewcommand{\arraystretch}{1.3}
\resizebox{\textwidth}{!}{%
\begin{tabular}{@{}c p{5cm} p{8.5cm}@{}}
\toprule
\textbf{Topic No} & \textbf{Theme and Associated Words} & \textbf{Example Posts/Comments} \\
\midrule
1 & \textbf{Vulvodynia} \newline test, symptom, doctor, infection, burning, cream, uti, vulvodynia, treatment, birth, bladder &
"I have been experiencing vestibulodynia for the past two years." \newline "She did conclude that this pain was at my vulva and mentioned vulvodynia" \\
\midrule
2 & \textbf{Endometriosis} \newline endo, surgery, endometriosis, period, doctor, cyst, symptom, surgeon, year, ultrasound &
"It is very scary to navigate endometriosis".  \newline "How frequently do they deal with cysts 10+ cm?"\\
\midrule

3 & \textbf{Impact of chronic pain} \newline work, life, care, understand, chronic, friend, hope, able, job, illness &
"Accepting that I was no longer able to be 100\% independent and self-sufficient was EXTREMELY difficult for me" \\
\midrule

4 & \textbf{IBS\& Diet} \newline eat, food, diet, ibs, stomach, drink, bad, trigger, food, help, sugar &
"I have IBS-M and had to cut out foods with fiber." \newline "I struggled with IBS for the past 10 years. Wherever I went, I had to make sure a bathroom was nearby." \\
\midrule

5 & \textbf{Sinusitis} \newline sinus, nose, nasal, infection, ent, allergy, rinse, eye, love, post &
"Sphenoid sinus infections can cause really brutal headaches since they’re so deep in your head." \newline "But that led me to a CT scan which showed all 8 sinuses infected and completely blocked." \\
\midrule

6 & \textbf{Pelvic \& lower back pain} \newline pelvic, nerve, muscle, floor, pt, therapy, exercise, adhesion, stretch, physical &
"Lower Back Pain - Ive had sciatic pain since I was around 18 years old " \\
\midrule

7 & \textbf{Social support/Emotional glue} \newline thank, share, good, great, appreciate, wow, glad, helpful, advice, awesome &
"I’m glad everyone has chimed in with their two cents." \newline "Thank you. I hope you get some relief from your pain as well." \\
\midrule

8 & \textbf{Migraine \& Headache} \newline migraine, headache, bad, stop, medication, sleep, med, fatigue, start, work &
"I’m new to this migraine rollercoaster. I’ve been struggling with what I now know are migraines recurrently" \\
\bottomrule
\end{tabular}%
}

\end{table}
\subsection{Gender and Drugs}

\begin{table}[!htb]
    \centering
     \caption{Efficacy and Emotion Distribution Percentages of Drugs by Gender}
    \label{tab:drug_effectiveness}
    \resizebox{\textwidth}{!}{
    \begin{tabular}{l l l c c c c c c c }
        \hline
        \textbf{Drug} & \textbf{Gender} & \shortstack{\textbf{Total} \\ \textbf{Mentions}} &\textbf{Effective (\%)} & \textbf{Ineffective (\%)} & \textbf{Side Effect (\%)} & \textbf{Anger (\%)} & \textbf{Fear (\%)} & \textbf{Joy (\%)} & \textbf{Sadness (\%)}\\
        \hline
        Amitriptyline & Female & 180 & \textbf{68.3} & 1.1 & 6.7 & 23.89 & \textbf{16.67} &43.33 &16.11 \\
                      & Male   & 6 & 66.7 & 0.0 & 0.0 & 16.67 & 33.33 &\textbf{33.33} &16.67\\
        Gabapentin    & Female & 827 & 66.7 & 0.8 & 6.4 & 15.24 &17.17 &41.72 &24.18\\
                      & Male  & 162 & 53.1 & 1.2 & 4.3 & 19.14 &19.75 &39.51 &20.37\\
        Lidocaine     & Female & 471 & 64.1 & 0.9 & 4.5 & 17.41 &16.14 &41.83 &22.51\\
                      & Male   & 81 & 59.3 & \textbf{3.7} & 7.4 & 13.58 &8.64 &46.91 &29.63\\
        Morphine      & Female & 325 & 49.8 & 1.2 & 6.2 & \textbf{22.77} &22.15 &31.38 &21.85\\
                      & Male   & 52 & 63.5 & 3.8 & 1.9 & 7.69 &13.46 &\textbf{55.77} &23.08\\
        Pregabalin    & Female & 435 & 58.2 & 1.4 & 7.1 & 18.85 &\textbf{16.75} &36.32 &26.90\\
                      & Male   & 111 & 56.8 & 0.9 & 4.5 & 11.71 &8.11 &\textbf{57.66} &20.72\\
        Tramadol      & Female & 296 & 57.1 & 1.0 & 6.4 & 20.61 &12.50 &41.22 &\textbf{23.65}\\
                      & Male   & 84 & 58.3 & 1.2 & 1.2 & 19.05 &\textbf{21.43} &46.43 &11.90\\
        \hline
    \end{tabular}
    }
   
\end{table}

The descriptive analysis of drug outcomes across genders is presented in Table \ref{tab:drug_effectiveness}. Several notable differences were observed in the proportion of individuals reporting drug effectiveness or experiencing side effects. To further investigate these patterns, we conducted Z-tests (with a significance threshold of 
p=0.05) on measures of efficacy and emotional responses. A detailed discussion of the findings for each drug is provided below.

\subsubsection{Amitriptyline}
In the case of efficacy, a statistically significant difference was observed between genders (p=0.04), with a greater proportion of females reporting the drug as effective. However, the p-values for medication ineffectiveness (p = 0.18) and side effects (p=0.086) were not statistically significant. Although previous studies did not explicitly confirm that amitriptyline is more effective for females, existing research has indicated that amitriptyline is generally safer for use in female patients \cite{b34}. Furthermore, findings from another study indicated a greater propensity for men to experience restless leg syndrome as a side effect compared to women \cite{b33}.

\subsubsection{Gabapentin}
Upon conducting the statistical analysis for gabapentin, no significant differences were observed in either efficacy or emotional responses between genders hence we did not investigate the experiences using this medication further.

\subsubsection{Lidocaine}
The statistical analysis revealed that a higher number of males reported lidocaine as being ineffective. No significant differences were observed for other efficacy outcomes or emotional responses. Previous studies have demonstrated mixed results regarding lidocaine’s effectiveness across genders \cite{b35}. One study found that intramuscular lidocaine infusion reduced pain intensity induced by hypertonic saline injection in men but not in women \cite{b74}. Conversely, another survey reported that women were more likely to respond to inhaled lidocaine than men in the treatment of cluster headaches \cite{b75}.

\subsubsection{Morphine}
Although no significant differences were observed in terms of efficacy, marked variations in emotional expression were evident between the genders. A significantly higher proportion of females expressed anger (p = 0.012), while males were more likely to express joy (p = 0.001). Previous research on Morphine and gender differences has indicated that Morphine is associated with a higher incidence of adverse effects nausea, dizziness in women compared to men \cite{b48}. This suggests a possible link between adverse reactions such as mood swings and respiratory depression and women's increased expression of anger with morphine use.

\subsubsection{Pregabalin}
For the drug Pregabalin, women showed a significant tendency to express fear (p = 0.02), whereas males exhibited a strong significance for expressing joy (p = 0.00). Although a preclinical study suggested that female mice may be more prone to pregabalin abuse and tolerance than male mice \cite{b37}, there is no direct evidence linking pregabalin abuse to fear responses, nor is there any confirmation of such findings in human populations.

\subsubsection{Tramadol}
Discussions on Reddit around Tramadol revealed interesting statistical patterns. Both males and females showed significant differences in expressing negative emotions: a higher number of males expressed fear (p = 0.04), while a higher number of females expressed sadness (p = 0.019). Upon reviewing previous research, we found compelling evidence supporting gender-specific adverse effects associated with Tramadol use. A study reported a higher risk for women to develop tramadol-associated vomiting compared to men among patients with chronic non-cancer pain \cite{b38}. Similarly, research conducted in the United States revealed that women reported tramadol-associated adverse reactions more frequently than men \cite{b39}. These findings underscore the biological differences in how Tramadol affects males and females, which may partly explain the emotional expressions observed in our analysis. \newline

The key findings from our study are as follows: \begin{itemize} 
\item We classified Reddit posts by gender through user-level aggregation, achieving an F1 score of 0.86. 
\item Female users were observed to express their pain experiences with greater emotional detail and were more likely to seek social support, whereas male users tended to adopt more solution-oriented approaches.
\item Conditions such as migraine, sinusitis, and vulvodynia were found to be more commonly reported among female users. 
\item Descriptive and statistical analyses of user-reported medication responses revealed that Lidocaine was associated with more negative outcomes among males, whereas Morphine and Pregabalin showed higher rates of negative impact among females. \end{itemize}

\section{Conclusion}
In this study, we used NLP to analyze Reddit posts on chronic pain, revealing gender-based differences in emotional expression and medication responses. These insights can enhance patient-provider communication, support demographic-specific care, and bring attention to important concerns such as medication side effects and inequities in care. By triangulating our data, these findings enhance the insights gained from clinical research. Our future work will involve collaborating with clinical specialists to examine how medications impact individuals of different genders, incorporating additional variables such as age to better understand the complexities and ultimately enable more targeted interventions.
\subsubsection{Acknowledgements} 
We would like to express our gratitude to Anna Tigunova for providing the RedDust dataset to support the development of our gender classifier.

%
%
%
%

\end{document}